\documentclass[11pt]{article}

\usepackage[margin=1in]{geometry}
\usepackage[T1]{fontenc}
\usepackage[utf8]{inputenc}
\usepackage{lmodern}
\usepackage{microtype}
\usepackage{graphicx}
\usepackage{booktabs}
\usepackage{longtable}
\usepackage{array}
\usepackage{calc}
\usepackage{amsmath,amssymb}
\usepackage{xcolor}
\usepackage{hyperref}
\usepackage{url}
\usepackage{enumitem}
\usepackage{float}
\usepackage{caption}
\usepackage{adjustbox}

\hypersetup{
    colorlinks=true,
    linkcolor=blue!50!black,
    citecolor=blue!50!black,
    urlcolor=blue!50!black,
    pdftitle={Under What Conditions Can a Machine Be Called Genuinely Creative?},
    pdfauthor={Yong Zeng}
}

\setlist[itemize]{topsep=0.25em,itemsep=0.15em,parsep=0pt}

\title{Under What Conditions Can a Machine Be Called \\Genuinely Creative?\\
\large A Designics-Based Requirement Framework for Creative Machine}
\author{Yong Zeng\\
Department of Cybersecurity and Intelligent Systems Engineering\\
Concordia University, Montreal, Canada\\
\texttt{yong.zeng@concordia.ca}}
\date{}

\begin{document}
\maketitle
\begin{abstract}
Recent artificial intelligence systems can generate texts, software architectures, hypotheses, designs, and scientific workflows that appear creative. These developments raise a foundational question that cannot be answered by output-centered benchmarks alone: under what conditions can a machine be called genuinely creative, and how can human agency be preserved within shared cognitive and creative environments? This paper addresses this question by developing a requirement framework derived from Designics, understood as the science of meaning-bearing intentional change.

The paper argues that genuine machine creativity should not be defined by output novelty, current performance, or transient technological architecture alone. Instead, creativity is understood as the structural transformation of incomplete situations through recursive intervention dynamics. Under this view, a machine can be called genuinely creative to the extent that it functionally and relationally participates in recursive intervention dynamics through an interdependent set of ten requirements: environment representation, scoped perception, conflict identification, intervention capability, consequence observation, knowledge and environment update, rescoping, local-to-global unfolding, value-based scoping, and human–AI co-living. These requirements are organized through the three laws of Designics: perception, conflict, and capability.

To illustrate the computational tractability of these requirements, the paper maps selected computational and empirical studies that partially operationalize them across cyber-physical domains, including recursive element extraction and autonomous mesh generation, and cyber-biological domains, including neurophysiological tracking, human capacity zone analysis, and dynamic human–machine workload reallocation. The paper then positions contemporary AI paradigms, including open-ended systems, automated discovery frameworks, self-modifying agents, foundation models, and agentic workflows, as pressure cases: they demonstrate powerful generative means, but do not by themselves establish genuine machine creativity.

Finally, the paper argues that proactive AI ethics is internal to a Designics account of genuine machine creativity rather than an after-the-fact compliance filter. Value-based scoping and human–AI co-living must shape how creative machines perceive environments, identify conflicts, select interventions, observe consequences, update knowledge, and rescope future action. The resulting framework keeps autonomous capability accountable to human agency, responsibility, relationship, and collective flourishing.
\end{abstract}

\noindent\textbf{Keywords:} Designics; Creative Machine; machine creativity; computational creativity;
recursive intervention dynamics; proactive AI ethics; human--AI
co-living; requirement framework; meaning-bearing intentional change.

\section{Introduction: From Creative Outputs to Genuine Machine
Creativity}\label{introduction-from-creative-outputs-to-genuine-machine-creativity}

Artificial intelligence systems increasingly generate outputs that appear creative. Large language models, multimodal generators, code models, scientific discovery systems, autonomous agents, and AI-supported design tools can produce texts, images, programs, hypotheses, experimental plans, design alternatives, and research proposals. Some of these outputs satisfy commonly used criteria of creativity, such as novelty, usefulness, value, and surprise (Boden,
2004; Ritchie, 2007; Runco \& Jaeger, 2012). These developments have renewed interest in computational creativity, machine discovery, artificial scientific reasoning, and human–AI co-creation (Ackerman, 2025; Bommasani et
al., 2021; Colton et al., 2011; Wiggins, 2006).

The philosophical problem addressed in this paper is not simply whether such systems produce impressive outputs. It is what would justify attributing creativity to a machine beyond evaluating those outputs. Product-based criteria can identify novelty, usefulness, value, or surprise, but they do not by themselves determine whether the system participates in the process through which a creative outcome becomes meaningful, consequence-bearing, and revisable within an environment. This problem has become culturally and philosophically urgent as creative machines move from specialist research communities into everyday artistic, musical, textual, and design practices. Recent human-centered discussions of creative AI emphasize both the promise of such systems to amplify human creativity and the need to approach them intentionally rather than through blind optimism, passive acceptance, efficiency, or profit alone (Ackerman, 2025). This paper shares that concern with human meaning, but addresses it at a different level: it asks what structural conditions would make the attribution of genuine creativity to a machine defensible.

Yet the ability to generate creative-looking outputs does not by itself answer a deeper question: under what conditions can a machine be called genuinely creative? This paper treats that question functionally and relationally rather than as a claim about machine consciousness, personhood, or human-like subjective intention. A system may generate a novel artifact without representing the environment in which the artifact matters. It may produce a useful proposal without identifying the conflict that makes the proposal necessary. It may generate surprising alternatives without observing consequences, revising knowledge, rescoping the problem, or participating responsibly in the human environment affected by its outputs. Genuine machine creativity therefore cannot be reduced to output generation alone.

This distinction is increasingly important because contemporary AI expands the means of human and machine action. It accelerates writing, coding, design exploration, data analysis, image generation, simulation, scientific hypothesis generation, and decision support. However, the expansion of means does not automatically provide meaning. Faster generation does not guarantee wiser judgment. Larger models do not guarantee responsible intervention. More candidate outputs do not guarantee deeper understanding of human life, social consequences, ecological boundaries, or ethical responsibilities. The AI age therefore intensifies a fundamental tension between means and meaning.

This paper starts from that tension. Technology is a means. AI is a means. Creative Machine, in its computational form, is also a means. The more fundamental question is how individuals, collectives, and machines can participate in meaning-bearing intentional change. This question belongs to Designics, understood here as the science of meaning-bearing intentional change (Zeng, 2026). Designics extends a long tradition of design research that treats design as a distinctive mode of artificial, reflective, and situated knowledge production (Cross, 1982; Schön, 1983;
Simon, 1996), while grounding intentional change in environment-centered analysis, conflict identification, and recursive transformation (Zeng \&
Cheng, 1991; Zeng, 2002, 2015). Designics asks how situations are perceived, how environments are scoped, how conflicts are identified, how interventions are generated, how consequences are observed, how knowledge is updated, and how future action is recursively redirected.

From the perspective of Designics, creativity is not merely the production of novel outputs. Creativity is a form of intentional change through which new structures, relations, actions, meanings, or possibilities become viable. Such creativity unfolds through recursive intervention dynamics. A situation is perceived under constraints. A scope is formed. Within that scope, conflicts, insufficiencies, blockages, instabilities, or unrealized possibilities are identified. An intervention is generated. The intervention produces consequences. These consequences change the environment, update knowledge, and require rescoping. Through repeated cycles of scoped perception, intervention, consequence observation, and rescoping, a first viable structure may emerge.

This view shifts the question of machine creativity. Instead of asking only whether a machine can generate outputs that humans judge to be novel or useful, we ask whether a machine can participate in recursive intervention dynamics. Can it represent an environment rather than merely process inputs? Can it identify conflicts rather than merely respond to prompts? Can it generate interventions rather than merely produce candidates? Can it observe consequences, update knowledge, rescope future perception, and support local decisions that recursively produce broader structure? Can it operate under ethical, social, ecological, legal, and meaning-based constraints? These questions define the requirement space for Creative Machine.

The purpose of this paper is therefore not to propose a fixed Creative Machine architecture. Architectures change with technological evolution. Large language models, reinforcement learning systems, world models, knowledge graphs, simulation tools, autonomous agents, active inference systems, symbolic reasoning systems, and hybrid architectures may all contribute to future Creative Machine implementations. But none of these technologies should define the concept of Creative Machine. A more stable foundation is needed: a requirement framework derived from the nature of genuine creativity itself.

This paper develops such a framework. Its primary contribution is a requirement framework for warranted attribution of machine creativity, with Creative Machine understood as the computational direction in which these requirements may be implemented. Designics functions as the derivational foundation of the framework, while proactive AI ethics and human–AI co-living enter as internal requirements rather than as a separate third agenda. The paper argues that, under this account, a machine can be called genuinely creative to the extent that it functionally participates in recursive intervention dynamics within a meaning-bearing environment. This participation depends on ten interdependent requirements: environment representation, scoped perception, conflict identification, intervention capability, consequence observation, knowledge and environment update, rescoping, local-to-global unfolding, value-based scoping, and human–AI co-living. These requirements are organized through the three laws of Designics: perception, conflict, and capability. They are not tied to one technological architecture. They provide a Designics-based foundation for evaluating, comparing, and guiding future Creative Machine architectures as technologies evolve.

This requirement-based approach also reframes the ethical problem. If a machine participates in genuine creativity under this account, it does not merely produce artifacts; it intervenes in human environments. Its outputs and actions may change how people think, decide, relate, learn, trust, create, and live. Therefore, AI ethics cannot remain an external filter applied after generation. Ethics must become part of the creative process itself. Ethical, legal, ecological, social, epistemic, and meaning-based constraints must shape perception, scoping, conflict identification, intervention selection, consequence evaluation, knowledge update, and rescoping. This is the basis for proactive AI ethics.

This paper positions contemporary AI developments as pressure cases rather than as definitions of machine creativity. Recent advances in open-ended systems, automated scientific discovery, self-modifying agents, foundation models, and agentic workflows demonstrate increasing autonomous capability. They make the question of genuine machine creativity urgent, but they do not by themselves resolve it. The deeper issue remains: under what conditions can these emerging capabilities be meaningfully scoped, responsibly guided, and integrated into human–AI co-living?

The central claim of the paper is as follows:

A machine can be called genuinely creative not merely by generating novel outputs, but to the extent that it satisfies the functional and relational requirements for participating in recursive intervention dynamics. These requirements are grounded in the three laws of Designics: perception, which determines what the machine can represent, include, observe, value, and rescope; conflict, which determines what calls for intervention; and capability, which determines what the machine can transform, learn from, unfold, and sustain in human–AI co-living.

The paper makes four contributions to this requirement-centered aim. First, it reframes machine creativity from output generation to recursive intervention dynamics. Second, it derives a stable requirement framework for Creative Machine from the three laws of Designics. Third, it uses selected cyber-physical and cyber-biological studies to demonstrate the operational tractability of these requirements, showing that they are computationally grounded without reducing Creative Machine to any single implementation. Fourth, it explains why proactive AI ethics and human–AI co-living belong inside the requirement framework: creative machines, if they participate in recursive intervention dynamics, intervene in meaning-bearing human environments.

The remainder of the paper is organized as follows. Section 2 introduces Designics and recursive intervention dynamics as the theoretical foundation. Section 3 clarifies what counts as genuine machine creativity beyond novel output generation. Section 4 develops the requirement framework for Creative Machine and organizes the ten requirements through the three laws of Designics. Section 5 provides computational illustrations of the requirement framework through cyber-physical and cyber-biological studies. Section 6 positions contemporary AI systems as pressure cases that motivate, but do not define, the framework. Section 7 develops proactive AI ethics and human–AI co-living as internal boundary conditions for responsible machine creativity. Section 8 discusses why architectures may change while requirements remain stable. Section 9 concludes the paper.

\section{Designics and Recursive Intervention
Dynamics}\label{designics-and-recursive-intervention-dynamics}

The requirement framework developed in this paper is grounded in
Designics. Designics is concerned with meaning-bearing intentional
change (Zeng, 2026). It studies how individuals, collectives, and
designed systems perceive situations, scope environments, identify
conflicts, generate interventions, observe consequences, update
knowledge, and redirect future action. In this sense, Designics is
broader than artifact design, engineering design, or technological
innovation. It builds on, but also extends, earlier accounts of design
as artificial science, reflective practice, designerly knowing, and
environment-based reasoning (Cross, 1982; Schön, 1983; Simon, 1996; Zeng
\& Cheng, 1991; Zeng, 2015). It treats design as a general structure of
intentional change in human life, social systems, and machine-supported
environments.

For machines, the term intentional change is used operationally rather
than psychologically. It does not assume subjective will, consciousness,
or human-like inward purpose. In Creative Machine, intention is
operationalized as goal-directed, scoped, consequence-sensitive
intervention within an environment. Such operational intentionality may
be externally assigned, system-maintained, recursively updated, or
embedded within a broader human-framed intentional process.

The central concern of Designics is not simply how something new is
produced, but how change becomes meaningful, responsible, and viable
within an environment. A design act is not only an internal act of
imagination or an external act of production; it is an intervention into
an environment. Such intervention changes conditions, relations,
resources, meanings, constraints, and future possibilities. Therefore,
the study of design must include not only the artifact or output, but
also the situation from which the intervention arises and the
consequences through which the environment is transformed.

This leads to the concept of recursive intervention dynamics. This
dynamic describes how intentional change unfolds through repeated
cycles. A conceptual schema of this cycle is as follows:

\begin{itemize}
\item
  Situation: the initial state of concern.
\item
  Constrained perception: the bounded intake of situational information.
\item
  Scoping: bounding the actionable environment.
\item
  Conflict identification: detecting insufficiencies, blockages, or
  instabilities.
\item
  Intervention: introducing an active change into the environment.
\item
  Consequence observation: evaluating the effects of the intervention.
\item
  Knowledge and environment update: integrating the observed changes.
\item
  Rescoping: adjusting the boundary for the next cycle.
\end{itemize}

This sequence should not be understood as a rigid linear algorithm. In
actual design and life processes, perception, scoping, conflict
identification, intervention, consequence evaluation, and knowledge
update often overlap. The sequence identifies conceptual dependencies
rather than temporal steps. Scope makes perception actionable.
Perception reveals conflicts or insufficiencies. Conflict motivates
intervention. Intervention produces consequences. Consequences change
the environment and the actor's knowledge. These changes require
rescoping, and rescoping makes the next cycle possible.

Recursive intervention dynamics becomes most visible when a situation is
incomplete, unstable, or underdetermined. In routine problem solving,
the problem, goal, knowledge, and evaluation criteria may be relatively
fixed. In creative situations, these elements are not stable in advance.
The problem may be unclear. The relevant environment may be only
partially perceived. The available knowledge may be insufficient. The
desired outcome may not yet be representable. Under such conditions,
creation cannot proceed by simple optimization within a predefined
space. It must proceed through recursive intervention.

The Environment plays a central role in this process. An environment is
not merely the external background of action; it is the structured field
within which intention becomes meaningful. It includes objects,
relations, resources, constraints, conditions, stakeholders, values,
conflicts, and possible consequences. Which parts of the environment
become active depends on scope. A technical scope may reveal one
conflict; a social scope may reveal another; an ethical scope may reveal
a hidden consequence; a temporal scope may reveal long-term instability.
Environment and scope must therefore be treated together.

Scoping is the act of selecting, bounding, and organizing the portion of
the environment that will be perceived, reasoned about, acted upon, and
evaluated. Scoping is necessary because no actor, human or machine, can
perceive the entire environment at once. Perception is always
constrained. This constraint is not merely a weakness; it is also a
condition for action. Without scope, perception becomes unbounded and
action becomes impossible. With a wrong scope, important conflicts may
remain hidden. With a rigid scope, the system may become trapped. With a
recursively revisable scope, however, the system can learn from
consequences and redirect future action.

Conflict is the trigger of intervention. In Designics, conflict should
not be reduced to formal contradiction. A conflict is an
environment-centered insufficiency, blockage, degradation, instability,
incompatibility, or unrealized possibility that prevents the current
situation from supporting a desired or viable structure. A conflict may
arise because a resource is missing, a condition is incompatible, an
interaction is blocked, a consequence is unacceptable, a value is
violated, or shadowed consequences produced by a previous intervention
have become active within the current scope. Conflict reveals where the
current environment is unable to sustain meaningful intentional change.

Intervention is the active mechanism of design. An intervention may add,
remove, separate, connect, weaken, strengthen, redirect, reorganize,
transform, or reinterpret objects, relations, conditions, resources, or
meanings within an environment. It may be physical, conceptual, social,
computational, institutional, ethical, or symbolic. An intervention is
not merely an output. It is a change introduced into an environment. Its
significance depends not only on what it produces immediately, but on
how it changes the conditions for future perception, action, and
meaning.

Consequence distinguishes recursive intervention dynamics from mere
generation. A generated output may be evaluated merely as an artifact.
An intervention, however, must be evaluated by its consequences.
Consequences may be intended or unintended, local or propagated,
immediate or delayed, constructive or destructive. A consequence may
resolve the original conflict, transform it, expose a deeper conflict,
or create a new one. Therefore, consequence observation is essential to
creativity, ethics, and learning. Without consequence observation, the
system cannot know whether its intervention has meaningfully changed the
environment.

Knowledge is also transformed through intervention. In routine
generation, knowledge may be treated as a fixed resource to be retrieved
and applied. In recursive intervention dynamics, knowledge is tested,
revised, expanded, and sometimes destabilized by consequences. A failed
intervention may reveal a hidden condition. An unexpected consequence
may reveal a new relation. A repeated pattern of failure may suggest a
new primitive law. A successful local intervention may become reusable
knowledge for future situations. Thus, knowledge is not only used in
design; it is also produced through design.

This view is directly grounded in Environment-Based Design. EBD treats
design as a process that begins from the environment, serves the
environment, and changes the environment (Zeng, 2004, 2015). Its core
methodological and mathematical logic of environment analysis, conflict
identification, and solution generation can be interpreted as an
important precursor and expression of recursive intervention dynamics
(Zeng \& Cheng, 1991; Zeng, 2002). Environment analysis corresponds to
scoped perception. Conflict identification reveals the insufficiency or
instability that motivates action. Solution generation becomes
intervention. The consequence of intervention updates the environment
and requires the next cycle of scoping.

Recursive intervention dynamics also explains why global creativity can
emerge from local action. A principled local intervention does not
remain isolated; it changes the environment for future interventions.
When each local intervention is scoped, consequence-sensitive, and
recursively updated, local decisions can produce emergent, highly
complex global structures. This principle is visible across design,
social behavior, and computational processes---ranging from rigid
geometric mesh generation to highly dynamic human-AI teaming loops. In
each case, global structure unfolds through a sequence of local
decisions whose consequences continuously reshape the next local
situation.

This local-to-global principle is critical for understanding both human
and machine creativity. A creative person does not usually create by
perceiving the entire future structure in advance. Rather, the person
acts within a scope, observes consequences, learns, adjusts, and
continues. Similarly, a genuinely creative machine should not be defined
by whether it can generate a complete final output in a single prompt.
It must be evaluated by whether it can participate in a recursive
process through which local interventions, consequence observation,
knowledge update, and rescoping lead toward a viable structure.

Recursive intervention dynamics therefore provides the theoretical
bridge from Designics to Creative Machine. Designics provides the
science of meaning-bearing intentional change. Recursive intervention
dynamics provides the mechanism through which such change unfolds.
Creative Machine, in this paper, names the computational direction
through which this mechanism may be supported, implemented, and
evaluated.

This distinction is important. A Creative Machine should not be defined
by a particular software architecture, large language model, multi-agent
framework, or learning algorithm. Such architectures will change with
technological development. What remains more stable are the requirements
derived from recursive intervention dynamics. Any machine that claims
genuine creativity must have the structural capacity to perceive a
scoped environment, identify conflicts, intervene, observe consequences,
update knowledge and environment, rescope future action, and participate
responsibly in human co-living.

Thus, the next question is not whether a machine can produce outputs
that appear creative, but what counts as genuine machine creativity when
evaluated through the lens of Designics. That question is addressed in
the next section.

\section{What Counts as Genuine Machine
Creativity?}\label{what-counts-as-genuine-machine-creativity}

The question of machine creativity is often approached through the
evaluation of outputs. A system is said to be creative if it produces
artifacts that are novel, useful, valuable, surprising, or aesthetically
interesting (Boden, 2004; Ritchie, 2007; Runco \& Jaeger, 2012; Wiggins,
2006). This output-centered view is important because creativity must
eventually appear in something: a text, an image, a design, a
hypothesis, or an action. However, output evaluation alone is
insufficient for defining genuine machine creativity.

This does not mean that creativity research has been limited to products
alone. A classic framework distinguishes four interrelated dimensions of
creativity: person, process, press, and product (Rhodes, 1961). The
point is rather that machine creativity is often operationalized through
product-like outputs: generated texts, images, programs, designs,
hypotheses, or benchmarks. From a Designics perspective, the other
dimensions must also be made computationally meaningful. The person
dimension corresponds to the creative agent's capability, knowledge,
affective condition, and capacity for action. The process dimension
corresponds to recursive intervention dynamics. The press dimension
corresponds to the environment, scope, constraints, resources, and
consequences within which creativity unfolds. The product dimension
corresponds to the viable structure or artifact that emerges from this
dynamic. Thus, the requirement framework proposed in this paper does not
reject product evaluation; it embeds product evaluation within a broader
recursive account of agent, process, and environment.

A machine may produce a novel output without participating in the
process through which that output becomes meaningful. It may generate an
unexpected image or hypothesis by recombining patterns from prior data.
It may produce useful alternatives because it has access to a massive
retrieval corpus or a large-scale optimization process. Such systems can
be highly productive and valuable. They may support human creativity.
Yet the deeper question remains whether the machine itself is creative
in a stronger sense, or whether it is primarily a powerful means for
producing candidate outputs.

This paper uses the term genuine machine creativity to refer to this
stronger condition, but it does so in a functional and relational sense.
The argument does not require the assumption that a machine possesses
human-like consciousness, subjective intention, or personhood. Rather, a
machine can be called genuinely creative to the extent that it
participates in the recursive process through which new viable
structures, relations, meanings, or actions become possible within a
meaning-bearing environment. Genuine creativity is not only the
production of a novel artifact; it is the capacity to transform the
conditions under which a new artifact, relation, explanation, or
intervention can emerge and become viable.

This distinction can be stated directly. Existing creativity research has commonly evaluated creativity through product qualities such as novelty, value, usefulness, and surprise; process accounts of generative, exploratory, and transformative activity; and agent- or system-level questions concerning autonomy, intention, understanding, and agency (Rhodes, 1961; Boden, 2004; Wiggins, 2006; Ritchie, 2007; Runco \& Jaeger, 2012; Jordanous, 2012). Against this background, the Designics-centered account adds a relational attribution criterion:

\begin{itemize}
\item \textbf{Product-based attribution} asks whether the machine produces something that appears novel, useful, valuable, or surprising.
\item \textbf{Process-based attribution} asks whether the machine participates in generative, exploratory, or transformative activity.
\item \textbf{Agent-based attribution} asks whether creativity depends on autonomy, intention, understanding, or agency.
\item \textbf{Designics-centered relational attribution} asks whether the machine functionally and relationally participates in recursive intervention dynamics through which new viable structures become possible within a meaning-bearing environment.
\end{itemize}

The proposed account is not intended to eliminate product-, process-, or agent-based criteria, but to show that they become philosophically stronger when embedded in a relational account of environment-transforming intervention. In this sense, the paper distinguishes three levels that are often conflated in discussions of machine creativity: creative output generation, where artifacts are judged as novel or valuable; creative process simulation, where systems model or enact mechanisms such as analogy, exploration, blending, or evaluation; and warranted attribution of genuine machine creativity, where the system participates in recursive, situated, value-scoped intervention.

This view is compatible with evaluation-oriented work in computational creativity. Jordanous's SPECS framework, for example, argues that creativity evaluation should begin by explicitly defining what creativity means in the target system and then deriving tests from that definition (Jordanous, 2012). The present paper follows that spirit at a more foundational level: it proposes recursive intervention dynamics as the basis for defining the conditions under which machine creativity can be attributed.

This distinction does not reject established computational creativity
criteria. Rather, it relocates them. Novelty, value, and surprise remain
important, but they become properties of interventions situated within
recursive environment transformation, rather than sufficient criteria
for genuine machine creativity by themselves (Boden, 2004; Colton et
al., 2011; Ritchie, 2007; Runco \& Jaeger, 2012; Wiggins, 2006).

This relocation also clarifies how to interpret historically important and contemporary creative AI systems. Symbolic systems such as AARON, music-generation systems such as EMI, evolutionary and rule-based systems, and more recent generative models can be important creative systems or creative tools without settling the stronger question of warranted attribution (Boden, 2004; Wiggins, 2006; Ritchie, 2007; Colton et al., 2011; Jordanous, 2012). Similarly, systems that explicitly reward style deviation, novelty, curiosity, or interestingness, such as Creative Adversarial Networks or curiosity-driven learning systems, internalize aspects of creative evaluation, but they still need to be assessed by how their generative capacities relate to environment, conflict, consequence, update, rescoping, and value (Elgammal et al., 2017; Schmidhuber, 2006).

This dynamic interpretation is also consistent with prior work on design
creativity as nonlinear design dynamics. Nguyen and Zeng (2012) proposed
that design reasoning follows nonlinear dynamics that may become chaotic
and that design creativity is related to mental stress through an
inverted-U relation. In that account, creative design emerges from
changing initial conditions involving design problems, design solutions,
design knowledge, and design-related information, while mental stress
depends on the relation between workload and mental capacity. The present
paper extends this line of reasoning from human design creativity to
machine creativity by asking what requirements a machine would need to satisfy in
order to participate in recursive intervention dynamics rather than
merely produce creative-looking outputs.

The second question is deeper because it concerns the process that gives
rise to creative emergence. From the perspective of Designics,
creativity involves meaning-bearing intentional change. A creative
system must therefore be evaluated by its participation in this dynamic,
not only by the surface properties of its outputs. This view yields six implications for evaluating a machine as genuinely creative under this account:

\begin{itemize}
\item \textbf{It requires an environment.} An output has meaning only within a context of use, interpretation, consequence, and value. A design proposal matters because it responds to an environment; a scientific hypothesis matters because it reorganizes observations within a field. A machine that produces outputs without representing or engaging the environment in which those outputs matter remains limited as a creative system.

\item \textbf{It requires constrained but revisable perception.} No creative actor perceives the whole environment at once. Creativity often depends on changing what is perceived as relevant. A machine becomes more genuinely creative when it can revise its scope of perception in response to conflict, consequence, failure, or new opportunity.

\item \textbf{It requires conflict recognition.} Creative action usually begins from some form of insufficiency: a blockage, a contradiction, a degraded structure, or an unrealized possibility. A machine that merely responds to user prompts may assist human creativity, but it does not independently identify what in the environment calls for intervention.

\item \textbf{It requires intervention and consequence observation.} Creativity requires action that changes the situation. What matters is that the machine introduces a change into the environment, and that the consequences of this change can be observed and evaluated. A candidate design may appear novel but fail when used. A machine that cannot observe or evaluate consequences cannot distinguish creative transformation from mere generation.

\item \textbf{It requires knowledge and environment update.} When consequences are observed, the system must learn from them. This learning may revise a model, expand primitive knowledge, expose a hidden relation, or reveal a new conflict. Genuine creativity requires that consequences not merely score outputs, but actively transform the system's future perception and action.

\item \textbf{It requires rescoping.} Rescoping is the revision of what should be perceived, what matters, what counts as conflict, and what future intervention is appropriate. Without rescoping, a system iterates within the same frame. With rescoping, it can change the frame. This is the key structural difference between routine optimization and genuine creativity.
\end{itemize}

These implications show why genuine machine creativity cannot be reduced
to a static measure of novelty. A random system can produce novelty. A
large generative model can produce novelty through statistical
recombination. An optimization system can produce novelty within a
predefined space. Genuine creativity requires novelty that emerges
through recursive intervention dynamics and becomes viable within a
scoped environment.

The concept of viability is critical. A creative outcome need not be
final, complete, or globally optimal. Many creative processes begin with
a partial structure that opens a new path. A first viable structure may
be an artifact, relation, explanation, category, or primitive knowledge
unit that becomes stable enough to support further development.
Viability means that the structure can function meaningfully within a
scope and can serve as a basis for further unfolding. This view is
consistent with design-theoretic accounts in which creative design
expands or transforms the space of possible concepts, knowledge, and
structures rather than merely selecting an optimal solution from a
predefined space (Hatchuel \& Weil, 2003, 2009; Zeng \& Gu, 1999a,
1999b).

This framework also distinguishes genuine creativity from mere autonomy.
A system may act autonomously without being creative---it may follow
goals, execute plans, call external tools, or optimize performance.
Genuine creativity requires more than autonomous execution; it requires
the ability to change the conditions of action. Autonomy concerns who or
what acts. Creativity concerns how new viable structure becomes possible
through action.

This distinction is essential for evaluating contemporary AI systems.
Foundation models, agentic workflows, and automated scientific discovery
systems may exhibit partial aspects of genuine machine creativity. Some
can generate novel artifacts or search large spaces. These developments
are important, but they should be evaluated according to stable
requirements rather than treated as final definitions of creativity. The
question is not whether a system appears impressive, but which
structural requirements for genuine creativity it satisfies, to what
degree, and which requirements remain underdeveloped.

The ethical implication is immediate. If a machine participates in
genuine creativity in this stronger sense, it is not merely producing
outputs for human
use; it is participating in intervention processes that reshape human
environments. Such creativity must therefore be guided by ethical,
social, ecological, legal, and meaning-based scoping. A creative machine
without value-based scoping may be powerful but dangerous. It may expand
means without disciplining meaning.

The argument of this section can now be summarized differently: genuine
machine creativity is not a property of an isolated output, but of a
system's participation in a recursive process that transforms incomplete
situations into viable structures. Novelty matters, but only when it
becomes situated, consequence-bearing, updateable, and value-oriented
within a scoped environment.

This definition prepares the requirement framework developed in the next
section. If genuine creativity requires participation in recursive
intervention dynamics, then the requirements for warranted attribution of machine creativity
should be structurally derived from the components of that dynamic. The
mathematical formalization of these requirements is reserved for future
work.

\section{Requirement Framework for Warranted Attribution of Machine
Creativity}\label{requirement-framework-for-creative-machine}

If genuine machine creativity is understood as participation in
recursive intervention dynamics, then a Creative Machine should not be
defined by a fixed technological architecture. Architectures change with
technological evolution. Large language models, reinforcement learning
systems, world models, agent frameworks, knowledge graphs, simulation
systems, active inference mechanisms, symbolic reasoning systems, and
hybrid architectures may all contribute to future implementations.
However, none of these technologies should define the concept of a
Creative Machine itself.

A more stable approach is to define requirements. Requirements specify what a machine would need to be able to do under this account of genuine machine creativity, regardless of how those capabilities are implemented. These requirements
are derived from the nature of recursive intervention dynamics:
constrained perception, scoping, conflict identification, intervention,
consequence observation, knowledge and environment update, rescoping,
local-to-global unfolding, and value-based orientation.

The framework should not be read as a sufficient engineering
specification or as a binary pass/fail checklist. The ten requirements
are proposed as necessary structural conditions under a Designics account
of genuine machine creativity, but they do not guarantee that any system
satisfying them superficially, weakly, or in a narrow domain is therefore
fully creative. They define the requirement space. Future work must
specify degrees of satisfaction, measurement criteria, domain-specific
implementations, and stronger formal conditions.

The ten requirements proposed in this section can be understood as an interdependent
system. Environment representation makes scoped perception possible;
scoped perception makes conflict identification possible; conflict
identification motivates intervention; intervention produces
consequences; consequence observation drives knowledge and environment
updates; updates require rescoping; and rescoping enables the next
cycle. Local-to-global unfolding explains how repeated scoped
interventions can produce emergent structure. Value-based scoping and
human--AI co-living ensure that machine creativity remains meaningful
and responsible within human environments.

\begin{figure}[t]
    \centering
    \includegraphics[
        width=1.03\textwidth
    ]{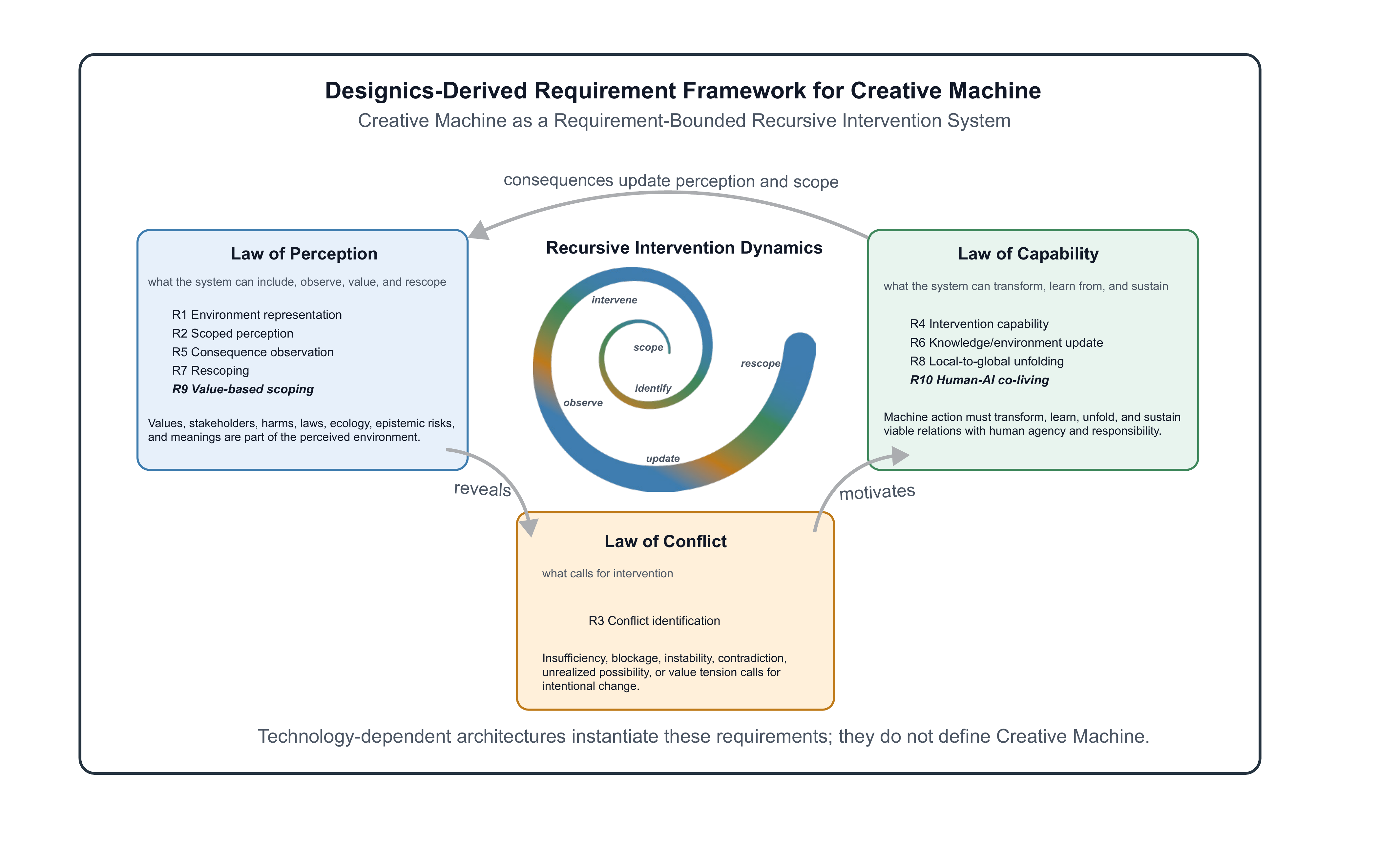}
    \caption{Designics-derived requirement framework for Creative Machine. The figure organizes the ten requirements through the three laws of Designics: perception, conflict, and capability. Perception grounds R1, R2, R5, R7, and R9; conflict grounds R3; and capability grounds R4, R6, R8, and R10. Together, these requirements form recursive intervention dynamics within the broader Designics framework of meaning-bearing intentional change. Technology-dependent architectures may instantiate these requirements, but they do not define Creative Machine.}
    \label{fig:framework}
\end{figure}

The ten requirements can also be understood through the three laws of
Designics: perception, conflict, and capability. The law of perception
concerns how an agent represents, scopes, observes, values, and rescopes
its environment; it grounds R1, R2, R5, R7, and R9. Under this view,
value is not an external ethical layer added after generation. Values,
stakeholders, harms, legal constraints, ecological boundaries, epistemic
risks, and meanings are part of the environment that must be perceived
and scoped. The law of conflict concerns how insufficiencies, blockages,
instabilities, contradictions, unrealized possibilities, and value
tensions become triggers for intentional change; it grounds R3. The law
of capability concerns how an agent transforms the environment, updates
knowledge, unfolds local interventions into broader structures, and
sustains viable human--AI co-living relations; it grounds R4, R6, R8,
and R10. This grouping shows that the proposed requirements are not an
arbitrary checklist. They are derived from the Designics account of how
meaning-bearing intentional change becomes possible. Figure~\ref{fig:framework}
summarizes this Designics-informed structure by showing how the three laws organize
the ten requirements within recursive intervention dynamics.

\subsection{R1: Environment Representation
Requirement}\label{r1-environment-representation-requirement}

A Creative Machine must operate within a represented environment, not
merely transform inputs into outputs. An environment is the structured
field in which creative action becomes meaningful. It includes objects,
relations, resources, constraints, conditions, stakeholders, values,
conflicts, possible interventions, and consequences. The machine must
represent enough of this field to perceive relevant objects, relations,
conflicts, constraints, and consequences, and this representation must
be updatable through intervention.

\subsection{R2: Scoped Perception
Requirement}\label{r2-scoped-perception-requirement}

A Creative Machine must perceive the environment through a bounded but
revisable scope. Scope defines what part of the environment is selected
for attention, reasoning, intervention, and evaluation. Because
unbounded perception produces computational overload, scope is
necessary. Because fixed scoping creates fixation, the machine must be
able to represent its own scope and revise it in response to conflict,
consequence, failure, or new opportunity.

\subsection{R3: Conflict Identification
Requirement}\label{r3-conflict-identification-requirement}

A Creative Machine must identify conflicts, insufficiencies, blockages,
instabilities, or unrealized possibilities within a scoped environment,
rather than merely receive them as human-specified tasks. Conflict is
the trigger of creative intervention. A machine that only responds to
explicit human instructions remains primarily a tool for human
creativity. A Creative Machine must participate in identifying what is
problematic, unstable, insufficient, or potentially transformable within
its environment, and must relate these conflicts to goals, values,
consequences, and future viability.

\subsection{R4: Intervention Capability
Requirement}\label{r4-intervention-requirement}

A Creative Machine must generate or enact interventions that transform
the scoped environment. An intervention is an intentional change
introduced into the environment: adding, removing, connecting,
redirecting, reorganizing, or reinterpreting objects, relations,
conditions, resources, or meanings. An output becomes an intervention
only when it is introduced into an environment and changes that
environment or its representation. A Creative Machine must therefore
produce changes that are environment-sensitive and consequence-bearing,
rather than merely producing candidate artifacts.

\subsection{R5: Consequence Observation
Requirement}\label{r5-consequence-observation-requirement}

A Creative Machine must observe and evaluate the consequences of its
interventions. Without consequence observation, a system cannot
distinguish between a meaningful intervention and a merely novel output.
It cannot identify unintended effects, learn from failure, or determine
whether a conflict has been resolved, transformed, deepened, or
displaced. A Creative Machine must evaluate consequences across its
relevant scope and, when needed, expand or shift that scope to capture
consequences that were initially hidden.

\subsection{R6: Knowledge and Environment Update
Requirement}\label{r6-knowledge-and-environment-update-requirement}

A Creative Machine must update both its knowledge and its environment
representation based on observed consequences. Creativity requires
learning from intervention. A consequence may confirm existing
knowledge, contradict it, expose a hidden condition, reveal a new
relation, or disclose a new conflict. The environment after intervention
is not the same as the environment before intervention. Therefore, the
machine must update the representation of the world in which the next
creative action will occur.

\subsection{R7: Rescoping Requirement}\label{r7-rescoping-requirement}

A Creative Machine must revise its boundary of perception, reasoning,
intervention, and evaluation in response to conflict, intervention, and
consequence. Rescoping is the recursive mechanism that differentiates
creativity from routine iteration. It allows the system to escape
fixation by shifting focus from technical objects to human practices,
from local performance to long-term stability, from artifact generation
to meaning-making, or from a narrow task frame to a broader environment
of consequence.

\subsection{R8: Local-to-Global Unfolding
Requirement}\label{r8-local-to-global-unfolding-requirement}

A Creative Machine must support principled local interventions that can
recursively produce emergent global structure. Creative structure does
not appear as a fully formed whole. It unfolds through sequences of
local decisions whose consequences reshape the next local situation. The
machine must participate in this unfolding, where each local action is
principled, consequence-sensitive, and recursively connected to future
possibilities.

\subsection{R9: Value-Based Scoping
Requirement}\label{r9-value-based-scoping-requirement}

A Creative Machine must include ethical, social, ecological, legal,
epistemic, and meaning-based constraints within the creative process
itself. Value-based scoping asks whose interests matter, what harms may
be hidden, what values are being served, and what forms of agency,
relationship, or meaning may be affected. Ethical scoping is a creative
requirement, not an external audit. It shapes what the machine
perceives, what conflicts it identifies, and what interventions it
considers viable.

\subsection{R10: Human--AI Co-Living
Requirement}\label{r10-humanai-co-living-requirement}

A Creative Machine must be evaluated by its effects on human co-living,
not only by output quality or task performance. If creativity is defined
only as output production, human--AI co-living may appear external to
creativity. If creativity is defined as intervention into
meaning-bearing environments, however, co-living becomes internal because
machine interventions can reshape the human environment in which meaning,
agency, responsibility, trust, and future creativity unfold. Therefore,
a Creative Machine must be evaluated by whether it supports human agency,
trustworthy knowledge, responsible intervention, sustainable cooperation,
and viable collective life. It must maintain a meaningful relationship
with its human partners rather than treating humans as passive consumers
of machine outputs.

\section{Computational Illustrations of the Requirement
Framework}\label{computational-illustrations-of-the-requirement-framework}

\subsection{The Purpose of the
Illustration}\label{the-purpose-of-the-illustration}

The requirement framework developed in this paper is not proposed as a
completed Creative Machine architecture. Instead, this section provides
a bounded illustration of the computational tractability of the proposed
requirements. It maps selected computational and empirical studies
across cyber-physical domains, including recursive element extraction
and autonomous mesh generation, and cyber-biological domains, including
neurophysiological tracking, human capacity zone analysis, and dynamic
human--machine workload reallocation (Jia et al., 2021; Kirgil-Budakli
et al., 2025; Pan et al., 2021, 2023; Yao et al., 2005; Zangeneh
Soroush \& Zeng, 2024; Zeng \& Cheng, 1991; Zeng, 2004, 2015; Zhao et
al., 2020, 2023, 2024).

The purpose is not to provide an exhaustive survey of computational
creativity, automated discovery, robotics, human--AI interaction, or
neuroergonomics. Rather, it is to show that the three-law structure of
Designics---perception, conflict, and capability---can be made
operational in bounded computational and human-integrated settings. The
cyber-physical and cyber-biological examples are therefore not treated
as separate lines of evidence, but as different domains in which the
same recursive intervention logic becomes visible.

The law of perception is illustrated by environment representation,
scoped perception, consequence observation, rescoping, and value-based
scoping. The law of conflict is illustrated by the identification of
insufficiencies, blockages, instabilities, and value tensions that
trigger intervention. The law of capability is illustrated by the
system's ability to intervene, update knowledge and environment, unfold
local actions into broader structures, and sustain viable human--AI
co-living relations.

To keep this illustration even-handed, the studies are treated not as
definitive solutions, but as structural illustrations. They are
positioned alongside broader developments in computational geometry,
reinforcement learning, human factors, adaptive automation, and
neuroergonomics. Their significance lies in showing that the Designics
requirements are not merely philosophical abstractions: they have
computational and empirical grounding across multiple domains. The
mathematical formalization of the full requirement framework is reserved
for future work.

\subsection{The Perception Law: R1, R2, R5, R7, and
R9}\label{the-perception-law-r1-r2-r5-r7-and-r9}

The law of perception concerns what the system can represent, scope,
observe, value, and rescope. In recursive mesh generation, perception is
operationalized through a represented geometric environment: domain
boundaries, local fronts, remaining unmeshed regions, quality
constraints, and topological conditions. The machine does not process an
unbounded mathematical universe. It bounds perception to a computable
local front, acts within that scope, observes the topological
consequence of each element extraction, and reconstructs the next scope
from the transformed boundary (Cheng \& Zeng, 1992; Zeng \& Cheng,
1993). This interpretation builds on broader traditions in computational
geometry and mesh generation, including advancing-front methods,
template-based meshing, and quality-driven discretization (Blacker \&
Stephenson, 1991; Owen, 1998).

The same perception law extends to cyber-biological environments. If a
machine is to co-live with humans, human cognitive capacity, stress,
attention, control, fatigue, and creative state become part of the
environment to be perceived. EEG microstate analysis in pilot-automation
contexts and EEG studies of idea generation, idea evolution, and
evaluation show that human cognitive states can be measured and
interpreted as part of the task environment (Jia \& Zeng, 2021; Zangeneh
Soroush \& Zeng, 2024; Zhao et al., 2020, 2024). These studies suggest
that future Creative Machine architectures may include
neurophysiological indicators within scoped perception, allowing
intervention to become sensitive to the human partner's cognitive state.

R9, value-based scoping, also belongs to the perception law. Values are
not external constraints added after technical generation; they are part
of the environment that must be perceived and scoped. Studies of
workload equilibrium and the human capacity zone show that human mental
capacity can be treated as a formal constraint on system action (Zhao et
al., 2023). In this sense, human well-being, agency, and sustainable
workload become perceptual and scoping conditions for machine
intervention. A machine that cannot perceive such value-bearing
environmental dimensions remains limited even if it can generate many
candidate outputs.

\subsection{The Conflict Law: R3}\label{the-conflict-law-r3}

The law of conflict concerns how insufficiencies, blockages,
instabilities, contradictions, unrealized possibilities, and value
tensions become triggers for intentional change. In recursive mesh
generation, the system detects unmeshable regions, collisions, invalid
topologies, distorted elements, or poor-quality configurations not
merely as terminal errors, but as computational conflicts (Zeng \& Yao,
2009). These conflicts indicate that the current local environment,
intervention rule, or scope is insufficient and that further action is
required.

In cyber-biological environments, conflict can appear as cognitive
overload, stress, degraded control, loss of agency, or a mismatch
between task demands and human capacity. Studies of network oscillations
during open-ended creation tasks show that high cognitive workload and
reduced cognitive control can occur during creative activity (Jia et
al., 2021). In a human--machine system, such cognitive overload can be
interpreted as an environmental conflict. Because R9 expands perception
to include values, stakeholders, human capacity, and meaning-bearing
relations, value tensions can also become conflicts that call for
intervention.

\subsection{The Capability Law: R4, R6, R8, and
R10}\label{the-capability-law-r4-r6-r8-and-r10}

The law of capability concerns what the system can transform, learn
from, unfold, and sustain. In mesh generation, the extraction or
generation of a single quadrilateral element constitutes a localized
intervention. Its immediate consequence is the mathematical and
topological update of the remaining boundary. The next decision is
therefore made in an environment partly created by the previous
intervention. The historical progression from rule-based element
extraction to artificial-neural-network-based extraction, self-learning
finite element extraction, and reinforcement-learning-based
quadrilateral mesh generation shows that primitive intervention rules
need not remain static heuristics (Pan et al., 2021, 2023; Yao et al.,
2005). They can be learned, adjusted, and improved based on the observed
consequences of prior topological interventions.

Capability also includes local-to-global unfolding. The FREEMESH
architecture showed that complex global mesh structures can arise from
recursive, consequence-sensitive local interventions (Zeng \& Cheng,
1993). The global structure does not need to be pre-calculated in its
entirety. It unfolds through local decisions whose consequences create
the next local decision context.

In cyber-biological environments, capability includes the ability to
sustain viable human--AI co-living. Dynamic workload reallocation for
human--robot teams shows how machine action can be evaluated not only by
task completion, but also by its effect on the human teammate
(Kirgil-Budakli et al., 2025). By measuring human stress or workload and
reallocating tasks accordingly, the system adjusts its intervention to
maintain a sustainable team relation. The human is not an external user
after the machine acts; the human is part of the shared environment in
which machine action must remain viable.

\subsection{Historical Mapping of the Designics Requirement
Framework}\label{historical-mapping-of-the-designics-requirement-framework}

Table~\ref{tab:historical-mapping} summarizes how selected
computational and cyber-biological studies illustrate the three-law
structure of the requirement framework. The table is not intended as an
exhaustive literature review. It identifies how different domains make
the same Designics structure operationally visible.

\begin{table}[H]
\centering
\caption{Historical mapping of Designics requirements across cyber-physical and cyber-biological work.}
\label{tab:historical-mapping}
\footnotesize
\begin{tabular}{@{}p{0.16\textwidth}p{0.16\textwidth}p{0.31\textwidth}p{0.31\textwidth}@{}}
\toprule
\textbf{Designics law} & \textbf{Requirements} & \textbf{Cyber-physical illustration} & \textbf{Cyber-biological illustration} \\
\midrule
Perception & R1, R2, R5, R7, R9 & Local geometric environment, boundary/front representation, consequence observation, and rescoping in recursive mesh generation & EEG indicators, workload/stress perception, human capacity zone, and value-based scoping of human well-being \\
Conflict & R3 & Unmeshable regions, invalid topologies, collisions, distorted elements, and poor-quality configurations & Cognitive overload, stress, degraded control, agency risk, and value tensions \\
Capability & R4, R6, R8, R10 & Element extraction, learned extraction policies, reinforcement learning, and local-to-global mesh unfolding & Workload reallocation, adaptive teaming, sustainable human--machine relation, and human--AI co-living \\
\bottomrule
\end{tabular}
\end{table}

\subsection{Summary}\label{summary}

The historical record summarized in this section shows that the
Designics requirements have computational grounding across both
cyber-physical and cyber-biological domains. In mesh generation,
recursive intervention dynamics appears as scoped geometric perception,
computational conflict, local intervention, consequence observation,
environment update, and global unfolding. In human-integrated systems,
the same logic extends to value-bearing perception, cognitive conflict,
workload redistribution, and sustainable human--machine relation.

These examples do not constitute a complete Creative Machine. Their
significance is more foundational. They show that the requirements
proposed in this paper are structurally viable and computationally
interpretable across domains. This gives the framework a stronger status
than a purely speculative theory.

The implication for contemporary AI is direct. Large language models,
multi-agent systems, automated discovery workflows, and self-modifying
agents provide powerful new means, but they often remain weakly grounded
in value-bearing perception, human cognitive consequence, and co-living.
The Designics requirement framework provides a way to evaluate these
systems not only by their generative power, but by whether their
interventions remain accountable to perception, conflict, and capability
within recursive intervention dynamics.

\section{Contemporary AI as Pressure
Cases}\label{contemporary-ai-as-pressure-cases}

Recent developments in open-ended systems, automated scientific
discovery, self-improving agents, foundation models, and agentic
workflows show that artificial intelligence is moving beyond one-shot
generation toward increasingly autonomous cycles of exploration,
production, evaluation, and modification (Bommasani et al., 2021; Lu et
al., 2024; Romera-Paredes et al., 2024; Shinn et al., 2023; Trinh et
al., 2024; Yao et al., 2023; Zhang et al., 2025). Earlier and parallel computational creativity work, including Creative Adversarial Networks, novelty search, and curiosity-oriented self-improvement, also shows how novelty, deviation, exploration, and evaluation can be partially internalized in computational systems (Elgammal et al., 2017; Lehman \& Stanley, 2011; Schmidhuber, 2006).

This section treats contemporary AI systems as pressure cases: cases
that challenge traditional concepts and force their underlying
structural requirements to become clearer. While these systems generate
outputs that appear creative, execute multi-step tasks, and modify parts
of their own operations, they do not by themselves establish genuine
machine creativity in the stronger sense proposed in this paper.

The central task is to evaluate these systems through the three
Designics laws: what they can perceive and scope, what they can identify
as conflict, what they can transform or sustain as capability, and what
limitations remain externally supplied or ethically underdeveloped.

\subsection{Analysis of Leading Pressure
Cases}\label{analysis-of-leading-pressure-cases}

Open-ended systems provide the first pressure case. Research on novelty
search, POET, and recent open-endedness frameworks emphasizes the
continuous production of artifacts, environments, or challenges that are
novel, learnable, and capable of supporting further adaptation (Hughes
et al., 2024; Lehman \& Stanley, 2011; Wang et al., 2019, 2020). This is
directly relevant to the idea of viable novelty. A system that
continually produces structures capable of supporting further learning
comes closer to creativity than a system optimizing a fixed objective.

However, open-endedness is insufficient for the Designics account of
genuine machine creativity. A system may produce novel, learnable
artifacts without adequately addressing value, meaning, human capacity,
or co-living. Open-ended exploration expands possibilities, but it does
not by itself discipline those possibilities by responsibility. It
becomes part of genuine creativity only when embedded in recursive
intervention dynamics that include environment representation,
consequence observation, rescoping, and value-based constraints.

Automated scientific discovery frameworks provide a second pressure
case. Recent AI-for-science and automated discovery systems attempt to
automate parts of the scientific research cycle, including idea
generation, literature search, code execution, analysis, hypothesis
refinement, and manuscript drafting (H. Wang et al., 2023; Lu et al.,
2024; Gottweis et al., 2025; Yamada et al., 2025). Related Nature
studies on mathematical and algorithmic discovery, such as FunSearch and
AlphaGeometry, further illustrate how foundation models can be coupled
with evaluators, symbolic engines, or search processes to produce
verifiable discoveries or proofs in constrained domains
(Romera-Paredes et al., 2024; Trinh et al., 2024). They move beyond
simple generation by participating in sequences resembling hypothesis
generation, intervention, consequence observation, and revision.

Nevertheless, these frameworks show why a stable requirement framework
remains necessary. Their research environments, evaluation criteria,
task domains, and benchmarks are often substantially provided by humans
or by pre-existing scientific infrastructure. Their consequences are
commonly measured through computational performance metrics,
experimental results, or simulated review rather than through broader
epistemic, social, ethical, or institutional effects. They generate
research products, but their status as genuinely creative systems depends
on the degree to which such products are integrated with the
meaning-bearing environment of science.

Self-improving and self-modifying agents provide a third pressure case.
Architectures designed to modify their own code, empirically validate
those modifications, and maintain an archive of evolving states
challenge the assumption that AI must operate within fixed boundaries
(Zhang et al., 2025; Schmidhuber, 2006; Zelikman et al., 2023). Related
embodied and open-ended agent systems further show how large language
models can support continual skill acquisition and environment
interaction (G. Wang et al., 2023).

Yet self-modification is not equivalent to genuine creativity. A system
may self-modify exclusively to optimize score metrics within a
predefined task environment. Its relevance to genuine machine creativity depends on whether
self-modification remains organized by human meaning, social consequence,
ecological constraint, and the psychological limits of human partners.
Without such organization, a system may become structurally more capable
without becoming more responsible.

Foundation models and agentic workflows provide a fourth pressure case.
Large language models and multimodal systems are powerful means for
semantic expansion, analogy, critique, hypothesis generation, code
generation, literature synthesis, and candidate generation (Bommasani et
al., 2021). Agentic workflows further connect these capabilities through
tool use, observation, memory, planning, and revision, as illustrated by
reasoning-and-acting frameworks, verbal reinforcement, tool-use
learning, and broader LLM-agent architectures (Shinn et al., 2023; Yao
et al., 2023; Schick et al., 2023; Xi et al., 2025). Together, they make
many components of an interactive machine technically plausible.

Nonetheless, foundation models and agentic workflows must be understood
as means, not definitions. Their creative status, however, depends on how their generative,
planning, tool-use, memory, and feedback capacities are organized. A
model may generate fluent outputs with limited environment
representation. An agentic workflow may repeat plan-act-observe cycles
while leaving rescoping externally supplied. A system may revise outputs
without updating primitive knowledge, or calculate confidence scores that
are not grounded in empirical measurement. Such systems become relevant
to Creative Machine to the degree that they are embedded within a
disciplined Designics ontology of environment, scope, conflict,
intervention, consequence, update, and value.

\subsection{Structural Comparison: Capability versus
Requirement}\label{structural-comparison-capability-versus-requirement}

Table~\ref{tab:ai-pressure-cases} summarizes how prominent contemporary AI paradigms can be evaluated through the three Designics laws. The table should not be read as a complete evaluation of any system; it indicates how selected pressure cases illustrate perception, conflict, and capability while also revealing remaining limitations.

\begin{table}[!htbp]
\centering
\caption{Contemporary AI pressure cases evaluated through the three Designics laws.}
\label{tab:ai-pressure-cases}
\scriptsize
\begin{tabular}{@{}p{0.17\textwidth}p{0.22\textwidth}p{0.17\textwidth}p{0.22\textwidth}p{0.16\textwidth}@{}}
\toprule
\textbf{AI pressure case} & \textbf{Perception law} & \textbf{Conflict law} & \textbf{Capability law} & \textbf{Main limitation} \\
\midrule
Open-ended systems & Perceive novelty and learnability within generated environments & Often weakly identify human or value-based conflicts & Generate adaptive structures and unfold possibilities & Value-based scoping and co-living remain underdeveloped \\
Automated scientific discovery & Represent literature, hypotheses, code, data, and results within bounded research environments & Identify research gaps or performance failures, often within human-supplied task frames & Generate hypotheses, run experiments, and revise outputs & Broader epistemic, institutional, ethical, and social consequences are weakly scoped \\
Self-modifying agents & Perceive performance limitations, code states, and archive states & Treat performance bottlenecks as conflicts & Modify code, validate changes, and archive improvements & Self-improvement may remain benchmark-centered \\
Foundation models and agentic workflows & Use context selection, retrieval, tool feedback, and memory as partial scoped perception & Conflicts are often prompt-defined or externally supplied & Provide strong generative and tool-use capability & Environment representation, rescoping, and value-based perception remain incomplete \\
\bottomrule
\end{tabular}
\end{table}
\subsection{Summary: Clarifying Two Interpretive Risks}\label{summary-clarifying-two-interpretive-risks}

The analysis of these pressure cases helps clarify two interpretive risks in
contemporary AI theory.

The first risk is dismissing contemporary AI as mere generation. Some
current systems clearly illustrate important partial requirements and
must be taken seriously as computational evidence for isolated
components of recursive intervention dynamics.

The second risk is treating contemporary AI capabilities as sufficient definitions of genuine creativity. Impressive engineering capabilities do not eliminate the
need for stable requirements derived from Designics.

The proper conclusion is that contemporary AI motivates the framework
without replacing it. Open-ended systems clarify the importance of
novelty and learnability (Lehman \& Stanley, 2011; Wang et al., 2019,
2020; Hughes et al., 2024). Automated scientific discovery systems
clarify the importance of hypothesis--intervention--consequence cycles
(H. Wang et al., 2023; Lu et al., 2024; Gottweis et al., 2025; Yamada et
al., 2025; Romera-Paredes et al., 2024; Trinh et al., 2024). Self-modifying agents clarify the importance of
architectural rescoping (Schmidhuber, 2006; Zelikman et al., 2023; Zhang
et al., 2025). Foundation models and agentic workflows clarify the
availability of powerful generative means (Bommasani et al., 2021; Yao
et al., 2023; Shinn et al., 2023; Schick et al., 2023; Xi et al., 2025).

Genuine machine creativity, however, requires more than these partial capacities. It demands that powerful means be disciplined by human
meaning, capacity, responsibility, and relationship.

This conclusion prepares the ethical framework developed in the next
section. If creative machines actively intervene in human environments
rather than merely deliver passive outputs, proactive AI ethics must
become an internal structural requirement of machine creativity rather
than an external filter applied after generation.

\section{Proactive AI Ethics and Human--AI
Co-Living}\label{proactive-ai-ethics-and-humanai-co-living}

This section clarifies the status of R9 and R10 within the requirement framework. It is not a separate theory of AI ethics added to the paper after the Creative Machine requirements have been developed. Rather, it explains why proactive AI ethics and human--AI co-living are internal requirements when machine creativity is understood as recursive intervention in meaning-bearing environments.

The three-law organization of the framework makes this internal relation explicit. Ethics enters first through the law of perception: values, stakeholders, harms, legal constraints, ecological boundaries, epistemic risks, and meanings are part of the environment to be perceived and scoped. Once values become part of the perceived environment, value tensions can become conflicts. Responsible intervention then becomes a capability requirement. Proactive AI ethics is therefore distributed across perception, conflict, and capability rather than appended as an external compliance filter.

This does not mean that every creative output must directly address social life. It means that machine creativity, when understood as intervention, changes the environments in which meaning, agency, responsibility, trust, learning, and future creativity unfold. A Creative Machine must therefore be evaluated not only by what it can generate, but by how its recursive dynamics affect human agency and shared life. This is consistent with recent calls for humble creative machines that elevate human capability rather than replace or dominate human creativity (Ackerman, 2025). The present framework gives this human-centered intuition a structural role through R9 and R10.

\subsection{The Mechanism of Proactive Ethical
Scoping}\label{the-mechanism-of-proactive-ethical-scoping}

A proactive approach to AI ethics begins with the systemic boundary
condition of scoping. Ethical scoping specifies what portion of the
environment the machine is permitted or required to consider. It
requires the system to account for whose interests are included, what
harms may be obscured, what systemic relationships are affected, and
what long-term vulnerabilities may be created by repeated interventions.
This orientation is consistent with value-sensitive and human-centered
approaches to technology design, which argue that human values,
stakeholders, and social consequences should shape design from the
beginning rather than appear only as after-the-fact evaluation criteria
(Friedman, Kahn, \& Borning, 2006; Shneiderman, 2020, 2022). In
recursive intervention dynamics, scoping ceases to be a purely technical
operation; it becomes a value-bearing perceptual act that shapes which
conflicts can be recognized and which interventions can be considered
viable.

This shift alters the core components of the Designics cycle.

Conflict identification is no longer only technical diagnosis. A
technically optimized intervention may be ethically unacceptable if it
degrades human agency, damages trust, or ignores vulnerable
stakeholders. Conversely, an ethical deficiency can itself act as the
primary conflict that triggers intervention. The machine must be able to
identify when a process reduces human accountability, creates unfair
access, or accelerates harmful cognitive offloading.

Intervention selection is no longer only a question of efficiency,
novelty, or statistical fluency. An intervention must be selected
according to its anticipated consequences within a value-scoped
environment. The machine must evaluate not only whether an action can be
done, but whether it should be done.

Consequence observation is no longer restricted to closed performance
benchmarks, user ratings, or accuracy scores. Machine-enacted
interventions may reshape how humans think, learn, communicate, trust,
decide, and relate. Therefore, consequence observation must include
human, social, ecological, legal, and epistemic dimensions.

This structural requirement addresses the core tension of the
contemporary technological landscape:

AI expands means; Designics disciplines meaning.

The rapid expansion of generative capability without a corresponding
discipline of meaning may induce systemic disorientation, dependency,
and loss of human agency. Proactive AI ethics places technology back in
its proper position: as a powerful means that must remain accountable to
human and collective flourishing.

\subsection{The Five Conflicts of Human--AI
Co-Living}\label{the-five-conflicts-of-humanai-co-living}

Human--AI co-living names the emerging condition in which humans and autonomous systems share cognitive, creative, and decision-making environments. Under an intervention-based account of creativity, this condition introduces at least five structural conflicts that a Creative Machine must be able to perceive and navigate.

First, there is a conflict of cognitive offloading: AI can assist inquiry, but extensive offloading may reshape attention, memory, task engagement, critical reasoning, and judgment (Risko \& Gilbert, 2016). Second, there is a conflict of agency: autonomous systems can increase output volume while weakening human control, responsibility, and supervisory engagement (Parasuraman, Sheridan, \& Wickens, 2000; Shneiderman, 2020). Third, there is a conflict of meaning: systems optimized for productivity may ignore purpose, identity, responsibility, trust, and shared understanding. Fourth, there is a conflict of social relation: autonomous interventions increasingly mediate communication, authorship, authority, accountability, and care. Fifth, there is a conflict of epistemic responsibility: foundation models and agentic systems can support knowledge work while also obscuring uncertainty, masking weak assumptions, or allowing fabrications to acquire false authority (Bommasani et al., 2021).

These conflicts are not external ethical topics. They are examples of value-bearing environmental conditions that affect what a Creative Machine should perceive, what it should treat as conflict, and which interventions can remain viable. In this respect, creative AI can be understood not only as a tool for producing artifacts, but also as a mirror that exposes human biases, conflicts of interest, failures of collaboration, and unresolved social tensions (Ackerman, 2025). Under a Designics account, these are not merely cultural observations; they are environmental conditions that must enter value-based scoping.

\subsection{Integration with the Requirement
Framework}\label{integration-with-the-requirement-framework}

To prevent proactive ethics from being reduced to a superficial add-on,
Table~\ref{tab:ethical-operationalization} shows how ethical and co-living constraints are structurally
embedded into each stage of the Designics requirement framework. This is
aligned with value-sensitive and human-centered AI approaches, but
extends them by treating value-based scoping and human--AI co-living as
internal requirements of machine creativity itself (Friedman et al.,
2006; Shneiderman, 2020, 2022).

\begin{longtable}[]{@{}
  >{\raggedright\arraybackslash}p{(\columnwidth - 4\tabcolsep) * \real{0.2027}}
  >{\raggedright\arraybackslash}p{(\columnwidth - 4\tabcolsep) * \real{0.2988}}
  >{\raggedright\arraybackslash}p{(\columnwidth - 4\tabcolsep) * \real{0.4985}}@{}}
\caption{\label{tab:ethical-operationalization}Ethical operationalization of the requirement framework.}\\
\toprule\noalign{}
\begin{minipage}[b]{\linewidth}\raggedright
\textbf{Requirement core}
\end{minipage} & \begin{minipage}[b]{\linewidth}\raggedright
\textbf{Technical interpretation: means}
\end{minipage} & \begin{minipage}[b]{\linewidth}\raggedright
\textbf{Proactive ethical operationalization: meaning}
\end{minipage} \\
\midrule\noalign{}
\endhead
\bottomrule\noalign{}
\endlastfoot
R1 and R2: representation and scope & Bounding the computational
context, state space, or environment of operation & Value-based scoping:
including human stakeholders, ecological constraints, legal boundaries,
and meaning-bearing relations within the field of concern \\
R3: conflict identification & Detecting performance degradation,
inconsistency, insufficiency, or structural constraint & Value-sensitive
diagnostics: recognizing systemic bias, erosion of human control, hidden
harm, and long-term vulnerability as triggers for action \\
R4 and R5: intervention and consequence & Executing tool actions and
observing immediate system responses & Shadow consequence auditing:
evaluating delayed, propagated, and non-technical impacts on human
agency, trust, cognitive health, and social relation \\
R6 and R7: update and rescoping & Adjusting internal states and shifting
active operational frames & Epistemic humility: preserving uncertainty,
isolating uncalibrated assertions, revising confidence, and expanding
scope when hidden harms or excluded stakeholders emerge \\
R9 and R10: values and co-living & Setting operating constraints and
maintaining interaction limits & Symbiotic viability: evaluating system
success also by whether its recursive dynamics support or undermine
human agency, relationship, responsibility, and collective
flourishing \\
\end{longtable}

\subsection{Summary}\label{summary-1}

This framework shifts the evaluation of creative systems away from
the isolated production of novel artifacts and toward life-world
transformation. The goal is not to build machines that maximize
creativity in isolation, but to understand and enforce the conditions
under which machine creativity can responsibly participate in human
co-living.

The central claim of this section is:

Genuine machine creativity requires proactive AI ethics because creative
machines intervene in human environments. Ethics must enter scoping,
conflict identification, intervention, consequence evaluation, knowledge
update, and rescoping, so that machine creativity remains structurally
accountable to human agency, responsibility, relationship, and
collective flourishing.

This operationalization completes the conceptual bridge of this paper.
Designics provides the foundational science of intentional change;
recursive intervention dynamics provides the execution mechanism; the
ten requirements define the functional and normative conditions; and proactive AI
ethics ensures that the resulting Creative Machine remains tethered to
human meaning.

\section{Architectures Change, Requirements
Remain}\label{architectures-change-requirements-remain}

The central contribution of this paper is a requirement framework, not a fixed architecture. This distinction is both strategic and theoretical. Large language models, multimodal systems, reinforcement learning frameworks, world models, knowledge graphs, agentic workflows, active inference mechanisms, symbolic systems, and hybrid architectures will continue to evolve (Bommasani et al., 2021; Friston, 2010; Friston et al., 2017; Yao et al., 2023; Shinn et al., 2023; Xi et al., 2025). If Creative Machine were defined by any one of these infrastructures, its theoretical validity would be tied to a transient phase of engineering history.

A requirement-based approach avoids this limitation. It defines stable conditions against which changing architectures can be evaluated when they are claimed to support genuine machine creativity. One implementation may realize scoped perception through attention, retrieval, and context-window management; another through symbolic environment models, probabilistic inference, simulation, or neurophysiological sensing. These implementations differ, but they can still be compared by the role they play within recursive intervention dynamics.

The hierarchy can therefore be stated compactly:

\begin{itemize}[leftmargin=1.6em, itemindent=0em, label={}, itemsep=0.35em]
    \item \textbf{Designics:} the derivational foundation, understood as the science of meaning-bearing intentional change.
    \item \textbf{Recursive intervention dynamics:} the operational mechanism through which such change unfolds.
    \item \textbf{R1--R10 requirement framework:} the stable structural conditions for Creative Machine under a Designics account.
    \item \textbf{Creative Machine architectures:} technology-dependent embodiments, including LLMs, agentic workflows, symbolic systems, reinforcement learning systems, simulation systems, and hybrid systems.
    \item \textbf{Applications:} domain-specific deployments, such as mesh generation, pilot teaming, education, scientific discovery, claim verification, research quality assessment, or social design.
\end{itemize}

This hierarchy keeps the paper's three identities properly ordered. The main object is the Creative Machine requirement framework. Designics provides the derivational foundation. Proactive AI ethics and human--AI co-living are internal requirements because Creative Machine, if realized, would intervene in meaning-bearing human environments.

The framework can guide future architectures in several ways. It can support architectural specification by showing which requirements are missing or weak; implementation comparison by evaluating functional roles rather than model size or software stack; evaluation discipline by extending metrics beyond accuracy, fluency, and novelty; multi-domain adaptation by preserving continuity across cyber-physical, cyber-biological, and social environments; and proactive governance by embedding R9 and R10 inside the architecture rather than after generation.

The framework also remains open to multiple formalizations. Active inference may formalize uncertainty, epistemic action, and belief updating (Friston, 2010; Friston et al., 2017; Parr, Pezzulo, \& Friston, 2022). C-K theory may clarify concept-space expansion (Hatchuel \& Weil, 2003, 2009). Category-theoretic, topological, or algebraic models may help formalize structural transformation. Reinforcement learning, simulation, and human factors models may operationalize local action, consequence observation, workload equilibrium, and cognitive capacity in bounded domains (Pan et al., 2021, 2023; Wickens, 2008; Zhao, Qiu, \& Zeng, 2023). No single formalism is treated as the exclusive mathematical foundation for Creative Machine; each may clarify part of the requirement space.

The answer to the question of genuine machine creativity therefore cannot be found in a purely technological race. Technical implementations will continue to evolve, but the requirements for genuine machine creativity and responsible human--AI co-living provide a more durable basis for judgment. The frontier is not defined only by what can be rapidly automated, but by how autonomous means can remain accountable to human meaning.

\section{Conclusion}\label{conclusion}

This paper began with a foundational question: under what conditions can
a machine become genuinely creative? This question is increasingly
urgent because contemporary artificial intelligence architectures can
already generate fluent and impressive texts, images, programs,
hypotheses, designs, and research proposals. Some systems coordinate
agentic workflows, execute code modifications, run external
experiments, evaluate generated outputs, and optimize performance. These
developments make the prospect of machine creativity more plausible, but
they also make its definition more difficult.

The core argument of this paper is that genuine machine creativity
should not be defined primarily by output novelty, transient software
architectures, or current state-of-the-art performance. Output novelty
is insufficient because a system may produce surprising artifacts
without participating in the structural process through which those
artifacts become meaningful. Architecture is insufficient because
technological frameworks are transient. Performance is insufficient
because an optimization routine may maximize benchmark scores while
remaining structurally blind to its consequences for the human
life-world.

Instead, this paper established a requirement framework derived from
Designics, understood as the science of meaning-bearing intentional
change. From this perspective, creativity is not an isolated act of
generation. It is the structural transformation of incomplete situations
through recursive intervention dynamics. This dynamic unfolds through
constrained perception, scoping, conflict identification, intervention,
consequence observation, knowledge and environment update, and
rescoping. Through this disciplined process, a first viable structure,
relation, explanation, or meaning can emerge and unfold.

\subsection{Summary of the Core
Contributions}\label{summary-of-the-core-contributions}

The paper makes four primary contributions to the development of a Creative Machine requirement framework.

First, it extends the generation paradigm. It shifts the evaluation of
machine creativity away from output-centered categories alone and toward
participation in recursive intervention dynamics. A machine can be called
genuinely creative to the extent that it can represent environments,
identify conflicts, generate interventions, observe consequences, update
knowledge, and revise its own scope of action.

Second, it derives ten functional requirements for Creative Machine.
These requirements, R1--R10, define necessary structural conditions
under a Designics account of genuine machine creativity, regardless of
the underlying technology stack. They do not constitute sufficient
engineering specifications or a binary checklist. They are organized
through the three laws of Designics: perception, conflict, and
capability. The law of perception grounds environment representation,
scoped perception, consequence observation, rescoping, and value-based
scoping. The law of conflict grounds conflict identification. The law of
capability grounds intervention, knowledge and environment update,
local-to-global unfolding, and human--AI co-living. This organization
shows that the requirements are not an arbitrary checklist, but a
Designics-informed account of how meaning-bearing intentional change
becomes possible.

Third, it draws on selected computational and cyber-biological studies.
The trajectory from recursive mesh generation in the cyber-physical loop
to EEG-informed human teaming, workload modeling, and human capacity
analysis in the cyber-biological loop shows that the requirement
framework is not merely speculative. It has computational grounding
across distinct domains. These examples do not prove the existence of a
complete Creative Machine, but they demonstrate that the core
requirements are operationally meaningful and tractable in bounded
settings.

Fourth, it integrates proactive AI ethics into the structure of machine
creativity. Ethics is not treated as a compliance filter added after
generation. It is part of the creative loop itself because values,
stakeholders, harms, legal constraints, ecological boundaries,
epistemic risks, and meanings are part of the environment to be
perceived and scoped. Once values become part of perception, value
tensions can become conflicts, and responsible intervention becomes a
capability requirement. Proactive AI ethics is therefore internal to
recursive intervention dynamics.

\subsection{The Structural Implication for Evolving
Architectures}\label{the-structural-implication-for-evolving-architectures}

By positioning contemporary AI advances, including open-ended systems,
automated scientific discovery frameworks, self-modifying agents,
foundation models, and agentic workflows, as pressure cases, this paper
maps the structural boundaries of the current frontier (Wang et al.,
2019, 2020; Lu et al., 2024; Romera-Paredes et al., 2024; Trinh et al.,
2024; Yamada et al., 2025; Zhang et al., 2025; Bommasani et al., 2021;
Yao et al., 2023; Shinn et al., 2023).

Yet these paradigms also reveal important limitations when evaluated
against the Designics requirement profile. They may rely heavily on
human-supplied evaluation criteria, operate with limited grounding in
human cognitive capacity, or lack internal mechanisms for value-based
scoping and long-term co-living consequences. They are therefore
important pressure cases, not definitions of genuine machine creativity.

The stable hierarchy developed in this paper keeps these technologies in
their proper place: as means, not definitions. This hierarchy is
anchored by the guiding principle:

\begin{quote}
AI expands means; Designics disciplines meaning.
\end{quote}

Creative Machine should not be pursued as a technological race or as a
rigid software architecture to be fixed once for all. It is a
computational vehicle for supporting meaning-bearing intentional change.
Its success must be judged by how it participates in human--AI
co-living, where humans and autonomous systems increasingly share
cognitive, creative, and decision-making environments.

\subsection{Horizons for Future
Research}\label{horizons-for-future-research}

To translate this framework into long-term engineering and academic
practice, future research can proceed along five primary axes.

First, the requirement framework can be formalized rigorously across
multiple technical disciplines. Active inference may help model belief
updating, uncertainty, epistemic action, and surprise (Friston, 2010;
Friston et al., 2017; Parr et al., 2022). Category-theoretic,
topological, or algebraic models may help formalize boundary
transformation and structural relation. Human factors models may help
define human capacity zones, workload equilibrium, stress, and cognitive
limits (Wickens, 2008; Zhao, Qiu, \& Zeng, 2023).

Second, domain-specific translations of the ten requirements can be
developed for engineering synthesis, autonomous scientific discovery,
life design, adaptive education, collaborative robotics, social design,
and AI ethics.

Third, reference system architectures can be built to test selected
combinations of requirements within bounded testbeds. Examples include
automated rescoping coupled with consequence auditing, conflict
identification coupled with value-based scoping, or human workload
sensing coupled with intervention adjustment.

Fourth, new evaluation metrics can be developed beyond scalar accuracy,
fluency, or novelty scores. Evaluation should measure structural scope
adjustment, conflict transformation, epistemic calibration, consequence
awareness, knowledge expansion, and effects on human agency and
co-living.

Fifth, cyber-biological integration can be further explored. Future
systems may need to account for human stress, workload equilibrium,
cognitive state, and creative phase in real time, not to control human
beings, but to protect human agency, responsibility, and cognitive
health within shared human--AI environments.

The final claim of the paper is this:

\begin{quote}
A machine can be called genuinely creative to the extent that its creativity
participates in recursive intervention dynamics and remains structurally
accountable to meaning-bearing human--AI co-living. Implementations and
software architectures will continue to change, but these functional
requirements provide a durable foundation for judging, guiding, and
developing Creative Machine.
\end{quote}

\section*{Acknowledgements}

The presented research is supported by the Natural Sciences and Engineering Research Council of Canada (NSERC) through its Discovery Grant.

\section*{References}

Ackerman, M. (2025). Creative Machines: AI, Art \& Us. Wiley.
\url{https://doi.org/10.1002/9781394321506}

Amodei, D., Olah, C., Steinhardt, J., Christiano, P., Schulman, J., \&
Mané, D. (2016). Concrete problems in AI safety. arXiv:1606.06565.

Blacker, T. D., \& Stephenson, M. B. (1991). Paving: A new approach to
automated quadrilateral mesh generation. International Journal for
Numerical Methods in Engineering, 32(4), 811--847.
\url{https://doi.org/10.1002/nme.1620320410}

Boden, M. A. (2004). The creative mind: Myths and mechanisms (2nd ed.).
Routledge.

Bommasani, R., Hudson, D. A., Adeli, E., Altman, R., Arora, S., et al.
(2021). On the opportunities and risks of foundation models.
arXiv:2108.07258.

Cheng, G. D., \& Zeng, Y. (1992). Strategies for automatic finite
element modeling. Computers \& Structures, 44(4), 905--909.
\url{https://doi.org/10.1016/0045-7949(92)90477-H}

Colton, S., Charnley, J., \& Pease, A. (2011). Computational creativity
theory: The FACE and IDEA descriptive models. In Proceedings of the 2nd
International Conference on Computational Creativity (ICCC 2011) (pp.
90--95).

Cross, N. (1982). Designerly ways of knowing. Design Studies, 3(4),
221--227. \url{https://doi.org/10.1016/0142-694X(82)90040-0}

Cross, N. (2006). Designerly ways of knowing. Springer.

Elgammal, A., Liu, B., Elhoseiny, M., \& Mazzone, M. (2017). CAN: Creative Adversarial Networks, generating ``art'' by learning about styles and deviating from style norms. In Proceedings of the 8th International Conference on Computational Creativity (ICCC 2017).

Friedman, B., Kahn, P. H., Jr., \& Borning, A. (2006). Value sensitive
design and information systems. In P. Zhang \& D. Galletta (Eds.),
Human-computer interaction in management information systems:
Foundations. M. E. Sharpe.

Friston, K. (2010). The free-energy principle: A unified brain theory?
Nature Reviews Neuroscience, 11(2), 127--138.
\url{https://doi.org/10.1038/nrn2787}

Friston, K., FitzGerald, T., Rigoli, F., Schwartenbeck, P., \& Pezzulo,
G. (2017). Active inference: A process theory. Neural Computation,
29(1), 1--49. \url{https://doi.org/10.1162/NECO_a_00912}

Gottweis, J., Weng, W.-H., Daryin, A., Tu, T., Palepu, A., et al.
(2025). Towards an AI co-scientist. arXiv:2502.18864.

Hatchuel, A., \& Weil, B. (2003). A new approach of innovative design:
An introduction to C-K theory. In Proceedings of the 14th International
Conference on Engineering Design (ICED'03) (pp. 109--124).

Hatchuel, A., \& Weil, B. (2009). C-K design theory: An advanced
formulation. Research in Engineering Design, 19(4), 181--192.
\url{https://doi.org/10.1007/s00163-008-0043-4}

Hughes, E., Dennis, M., Parker-Holder, J., Behbahani, F., Mavalankar,
A., Shi, Y., Schaul, T., \& Rocktäschel, T. (2024). Open-endedness is
essential for artificial superhuman intelligence. In Proceedings of the
41st International Conference on Machine Learning, PMLR, 235,
20597--20616. arXiv:2406.04268.

Jia, W., von Wegner, F., Zhao, M., \& Zeng, Y. (2021). Network
oscillations imply the highest cognitive workload and lowest cognitive
control during idea generation in open-ended creation tasks. Scientific
Reports, 11, 24277. \url{https://doi.org/10.1038/s41598-021-03577-1}

Jia, W., \& Zeng, Y. (2021). EEG signals respond differently to idea
generation, idea evolution and evaluation in a loosely controlled
creativity experiment. Scientific Reports, 11, 2119.
\url{https://doi.org/10.1038/s41598-021-81655-0}

Jordanous, A. (2012). A standardised procedure for evaluating creative systems: Computational creativity evaluation based on what it is to be creative. Cognitive Computation, 4(3), 246--279.
\url{https://doi.org/10.1007/s12559-012-9156-1}

Kirgil-Budakli, R., Zeng, Y., \& Akgunduz, A. (2025). Dynamic workload
reallocation for human-robot teams based on real-time stress analysis.
AI EDAM, 39, e18. \url{https://doi.org/10.1017/S0890060425100073}

Lehman, J., \& Stanley, K. O. (2011). Abandoning objectives: Evolution
through the search for novelty alone. Evolutionary Computation, 19(2),
189--223. \url{https://doi.org/10.1162/EVCO_a_00025}

Lu, C., Lu, C., Lange, R. T., Foerster, J., Clune, J., \& Ha, D. (2024).
The AI Scientist: Towards fully automated open-ended scientific
discovery. arXiv:2408.06292.

Nguyen, T. A., \& Zeng, Y. (2012). A theoretical model of design
creativity: Nonlinear design dynamics and mental stress-creativity
relation. Journal of Integrated Design and Process Science, 16(3),
65--88. \url{https://doi.org/10.3233/jid-2012-0007}

Owen, S. J. (1998). A survey of unstructured mesh generation technology.
In Proceedings of the 7th International Meshing Roundtable.

Pan, J., Huang, J., Wang, Y., Cheng, G., \& Zeng, Y. (2021). A
self-learning finite element extraction system based on reinforcement
learning. AI EDAM, 35(2), 180--208.
\url{https://doi.org/10.1017/S089006042100007X}

Pan, J., Huang, J., Cheng, G., \& Zeng, Y. (2023). Reinforcement
learning for automatic quadrilateral mesh generation: A soft
actor-critic approach. Neural Networks, 157, 288--304.
\url{https://doi.org/10.1016/j.neunet.2022.10.022}

Parasuraman, R., Sheridan, T. B., \& Wickens, C. D. (2000). A model for
types and levels of human interaction with automation. IEEE Transactions
on Systems, Man, and Cybernetics---Part A: Systems and Humans, 30(3),
286--297. \url{https://doi.org/10.1109/3468.844354}

Parr, T., Pezzulo, G., \& Friston, K. J. (2022). Active inference: The
free energy principle in mind, brain, and behavior. MIT Press.
\url{https://doi.org/10.7551/mitpress/12441.001.0001}

Romera-Paredes, B., Barekatain, M., Novikov, A., Balog, M., Kumar, M. P.,
Dupont, E., Ruiz, F. J. R., Ellenberg, J. S., Wang, P., Fawzi, O., Kohli,
P., \& Fawzi, A. (2024). Mathematical discoveries from program search
with large language models. Nature, 625, 468--475.
\url{https://doi.org/10.1038/s41586-023-06924-6}

Risko, E. F., \& Gilbert, S. J. (2016). Cognitive offloading. Trends in
Cognitive Sciences, 20(9), 676--688.
\url{https://doi.org/10.1016/j.tics.2016.07.002}

Rhodes, M. (1961). An analysis of creativity. Phi Delta Kappan, 42(7),
305--310.

Ritchie, G. (2007). Some empirical criteria for attributing creativity
to a computer program. Minds and Machines, 17(1), 67--99.
\url{https://doi.org/10.1007/s11023-007-9066-2}

Runco, M. A., \& Jaeger, G. J. (2012). The standard definition of
creativity. Creativity Research Journal, 24(1), 92--96.
\url{https://doi.org/10.1080/10400419.2012.650092}

Schick, T., Dwivedi-Yu, J., Dessì, R., Raileanu, R., Lomeli, M.,
Zettlemoyer, L., Cancedda, N., \& Scialom, T. (2023). Toolformer:
Language models can teach themselves to use tools. Advances in Neural
Information Processing Systems, 36. arXiv:2302.04761.

Schmidhuber, J. (2006). Gödel machines: Fully self-referential optimal
universal self-improvers. In B. Goertzel \& C. Pennachin (Eds.),
Artificial general intelligence (pp. 199--226). Springer.

Schön, D. A. (1983). The reflective practitioner: How professionals
think in action. Basic Books.

Shinn, N., Cassano, F., Berman, E., Gopinath, A., Narasimhan, K., \&
Yao, S. (2023). Reflexion: Language agents with verbal reinforcement
learning. Advances in Neural Information Processing Systems, 36.
arXiv:2303.11366.

Shneiderman, B. (2020). Human-centered artificial intelligence:
Reliable, safe \& trustworthy. International Journal of Human--Computer
Interaction, 36(6), 495--504.
\url{https://doi.org/10.1080/10447318.2020.1741118}

Shneiderman, B. (2022). Human-centered AI. Oxford University Press.
\url{https://doi.org/10.1093/oso/9780192845290.001.0001}

Simon, H. A. (1996). The sciences of the artificial (3rd ed.). MIT
Press.

Trinh, T. H., Wu, Y., Le, Q. V., He, H., \& Luong, T. (2024). Solving
olympiad geometry without human demonstrations. Nature, 625, 476--482.
\url{https://doi.org/10.1038/s41586-023-06747-5}

Wang, G., Xie, Y., Jiang, Y., Mandlekar, A., Xiao, C., Zhu, Y., Fan, L.,
\& Anandkumar, A. (2023). Voyager: An open-ended embodied agent with
large language models. arXiv:2305.16291.

Wang, H., Fu, T., Du, Y., Gao, W., Huang, K., Liu, Z., et al. (2023).
Scientific discovery in the age of artificial intelligence. Nature,
620(7972), 47--60. \url{https://doi.org/10.1038/s41586-023-06221-2}

Wang, R., Lehman, J., Clune, J., \& Stanley, K. O. (2019). Paired
open-ended trailblazer (POET): Endlessly generating increasingly complex
and diverse learning environments and their solutions. arXiv:1901.01753.

Wang, R., Lehman, J., Rawal, A., Zhi, J., Li, Y., Clune, J., \& Stanley,
K. O. (2020). Enhanced POET: Open-ended reinforcement learning through
unbounded invention of learning challenges and their solutions. In
Proceedings of the 37th International Conference on Machine Learning.
arXiv:2003.08536.

Wickens, C. D. (2008). Multiple resources and mental workload. Human
Factors, 50(3), 449--455. \url{https://doi.org/10.1518/001872008X288394}

Wiggins, G. A. (2006). A preliminary framework for description, analysis
and comparison of creative systems. Knowledge-Based Systems, 19(7),
449--458. \url{https://doi.org/10.1016/j.knosys.2006.04.009}

Xi, Z., Chen, W., Guo, X., He, W., Ding, Y., Hong, B., et al. (2025).
The rise and potential of large language model based agents: A survey.
Science China Information Sciences, 68(2), 121101.
\url{https://doi.org/10.1007/s11432-024-4222-0}

Yamada, Y., Lange, R. T., Lu, C., Hu, S., Lu, C., Foerster, J., Clune,
J., \& Ha, D. (2025). The AI Scientist-v2: Workshop-level automated
scientific discovery via agentic tree search. arXiv:2504.08066.

Yao, S., Yan, B., Chen, B., \& Zeng, Y. (2005). An ANN-based element
extraction method for automatic mesh generation. Expert Systems with
Applications, 29(1), 193--206.
\url{https://doi.org/10.1016/j.eswa.2005.01.019}

Yao, S., Zhao, J., Yu, D., Du, N., Shafran, I., Narasimhan, K., \& Cao,
Y. (2023). ReAct: Synergizing reasoning and acting in language models.
In International Conference on Learning Representations.
arXiv:2210.03629.

Zangeneh Soroush, M., \& Zeng, Y. (2024). EEG-based study of design
creativity: A review on research design, experiments, and analysis.
Frontiers in Behavioral Neuroscience, 18, 1331396.
\url{https://doi.org/10.3389/fnbeh.2024.1331396}

Zelikman, E., Lorch, E., Mackey, L., \& Kalai, A. T. (2023). Self-taught
optimizer (STOP): Recursively self-improving code generation.
arXiv:2310.02304.

Zeng, Y. (2002). Axiomatic theory of design modeling. Transactions of
the SDPS: Journal of Integrated Design and Process Science, 6(3), 1--28.

Zeng, Y. (2004). Environment-based formulation of design problem.
Transactions of the SDPS: Journal of Integrated Design and Process
Science, 8(4), 45--63.

Zeng, Y., \& Yao, S. (2009). Understanding design activities through
computer simulation. Advanced Engineering Informatics, 23(3), 294--308.
\url{https://doi.org/10.1016/j.aei.2009.02.001}

Zeng, Y. (2015). Environment-Based Design (EBD): A methodology for
transdisciplinary design. Journal of Integrated Design and Process
Science, 19(1), 5--24. \url{https://doi.org/10.3233/jid-2015-0004}

Zeng, Y. (2026). Designics: A science of intentional change. Journal of
Integrated Design and Process Science. Advance online publication.
\url{https://doi.org/10.1177/10920617261450751}

Zeng, Y., \& Cheng, G. D. (1991). On the logic of design. Design
Studies, 12(3), 137--141.
\url{https://doi.org/10.1016/0142-694X(91)90022-O}

Zeng, Y., \& Cheng, G. D. (1993). Knowledge-based free mesh generation
of quadrilateral elements in two-dimensional domains. Microcomputers in
Civil Engineering, 8(4), 259--270.
\url{https://doi.org/10.1111/j.1467-8667.1993.tb00211.x}

Zeng, Y., \& Gu, P. (1999a). A science-based approach to product design
theory. Part I: Formulation and formalization of design process.
Robotics and Computer-Integrated Manufacturing, 15(4), 331--339.
\url{https://doi.org/10.1016/S0736-5845(99)00028-9}

Zeng, Y., \& Gu, P. (1999b). A science-based approach to product design
theory. Part II: Formulation of design requirements and products.
Robotics and Computer-Integrated Manufacturing, 15(4), 341--352.
\url{https://doi.org/10.1016/S0736-5845(99)00029-0}

Zhang, J., Hu, S., Lu, C., Lange, R., \& Clune, J. (2025). Darwin Gödel
Machine: Open-ended evolution of self-improving agents.
arXiv:2505.22954.

Zhao, M., Qiu, D., \& Zeng, Y. (2023). How much workload is a ``good''
workload for human beings to meet the deadline: Human capacity zone and
workload equilibrium. Journal of Engineering Design, 34(8), 644--673.
\url{https://doi.org/10.1080/09544828.2023.2249216}

Zhao, M., Jia, W., Yang, D., Nguyen, P., Nguyen, T. A., \& Zeng, Y. (2020). A tEEG framework for studying designer's
cognitive and affective states. Design Science, 6, e29.
\url{https://doi.org/10.1017/dsj.2020.28}

Zhao, M., Jia, W., Jennings, S., Law, A., Bourgon, A., Su, C., Larose, M.-H., Grenier, H., Bowness, D., \& Zeng, Y. (2024). Monitoring pilot trainees' cognitive control
under a simulator-based training process with EEG microstate analysis.
Scientific Reports, 14, 24632.
\url{https://doi.org/10.1038/s41598-024-76046-0}

\end{document}